\documentclass{article}

\usepackage[final]{neurips_2024}

\usepackage[utf8]{inputenc} % allow utf-8 input
\usepackage[T1]{fontenc}    % use 8-bit T1 fonts
\usepackage{booktabs}       % professional-quality tables
\usepackage{amsfonts}       % blackboard math symbols
\usepackage{nicefrac}       % compact symbols for 1/2, etc.
\usepackage{microtype}      % microtypography
\usepackage{xcolor}         % colors
 \usepackage{graphicx} 
 \usepackage{amsmath}
 \usepackage{times}  % DO NOT CHANGE THIS
\usepackage{helvet}  % DO NOT CHANGE THIS
\usepackage{courier}  % DO NOT CHANGE THIS
\usepackage[hyphens]{url}  % DO NOT CHANGE THIS
\usepackage{graphicx} % DO NOT CHANGE THIS
\usepackage{natbib}  % DO NOT CHANGE THIS AND DO NOT ADD ANY OPTIONS TO IT
\usepackage{caption} % DO NOT CHANGE THIS AND DO NOT ADD ANY OPTIONS TO IT
\usepackage{amsmath, amssymb, amsthm}
\usepackage{tikz}
\usetikzlibrary{arrows.meta, positioning}
\usepackage{booktabs}       % professional-quality tables
\usepackage{amsfonts}       % blackboard math symbols
\usepackage{nicefrac}       % compact symbols for 1/2, etc.
\usepackage{microtype}      % microtypography
\usepackage{xcolor}         % colors

\usepackage{booktabs} % Add this in the preamble if it's not already included

\usepackage{algorithm}
\usepackage{algorithmic}

\usepackage{newfloat}
\usepackage{listings}
\DeclareCaptionStyle{ruled}{labelfont=normalfont,labelsep=colon,strut=off} % DO NOT CHANGE THIS
\floatstyle{ruled}
\newfloat{listing}{tb}{lst}{}
\floatname{listing}{Listing}

\title{The Shepherd Test: How Will Super Intelligent Agents Balance Care and Control in Asymmetric Relationships?}

\author{%
  Djallel Bouneffouf, Matthew Riemer, and Kush R. Varshney 
    \\
  IBM Research\\
  1101 Kitchawan Rd, Yorktown Heights, NY 10598\\
  \texttt{Firstname.lastname@ibm.com} \\
}

\begin{document}

\maketitle

\begin{abstract}
This paper introduces the Shepherd Test, a new conceptual test for assessing the moral and relational dimensions of superintelligent artificial agents. The test is inspired by human interactions with animals, where ethical considerations about care, manipulation, and consumption arise in contexts of asymmetric power and self-preservation. We argue that AI crosses an important, and potentially dangerous, threshold of intelligence when it exhibits the ability to manipulate, nurture, and instrumentally use less intelligent agents, while also managing its own survival and expansion goals. This includes the ability to weigh moral trade-offs between self-interest and the well-being of subordinate agents. The Shepherd Test thus challenges traditional AI evaluation paradigms by emphasizing moral agency, hierarchical behavior, and complex decision-making under existential stakes. We argue that this shift is critical for advancing AI governance, particularly as AI systems become increasingly integrated into multi-agent environments. We conclude by identifying key research directions, including the development of simulation environments for testing moral behavior in AI, and the mathematical formalization of ethical manipulation within multi-agent systems.
\end{abstract}

\section{Introduction}

As artificial intelligence (AI) systems become more capable, the focus of alignment research has been to ensure that these systems remain beneficial and responsive to human goals \citep{russell2019human, christiano2018deep}. However, much of this work presupposes a one-sided relationship in which AI is subordinate to human oversight. In this paper, we ask a deeper and less explored question: What does it mean for an AI agent to be so intelligent that it begins to relate to other systems the way humans relate to animals? We propose that the moral asymmetry between humans and animals offers a revealing model for evaluating superintelligent AI.

Humans interact with animals through domestication, experimentation, companionship, and consumption. These relationships are complex: they involve care and control, nurturing and instrumentalization, moral concern and exploitation. Crucially, they are grounded in an asymmetry of intelligence, agency, and power. Animals are often treated as beings of lesser moral standing, sometimes deserving of ethical treatment but rarely considered equals \cite{cave2020problem}. The ethical frameworks governing human–animal relationships—such as those articulated in animal rights and utilitarian ethics—do not provide a blueprint for how AI should treat humans. Instead, they offer cautionary insights into how powerful beings often rationalize their treatment of weaker ones, highlighting the risks we face if superintelligent AI were ever to adopt a similar stance toward us \citep{singer1975animal, regan1983case}.

We introduce the Shepherd Test as a safeguard for evaluating whether an AI system has advanced to a form of general intelligence that includes the ability to engage with, and morally reason about, weaker agents. Much like humans design farms, pet-care routines, and scientific experiments around animals, a superintelligent AI may one day construct and manage simpler agents—artificial or biological—within its environment. The way such an AI chooses to interact with less capable entities offers a critical diagnostic: does it grasp ethical asymmetry, power dynamics, and moral responsibility, or does it risk exploiting them?

The test’s name is drawn from the pastoral role of a shepherd charged with ensuring the flock’s welfare (ethical concern), defending it from threats (competence), and at times making difficult sacrifices (tragic trade-offs). This metaphor underscores that superintelligence entails not just power but also responsibility. By framing the evaluation in this way, the Shepherd Test acts as a safeguard: a tool to detect whether an AI recognizes and respects the ethical boundaries that prevent dominance from turning into tyranny.

Critically, the Shepherd Test is not merely about measuring an AI’s attributes (e.g., raw cognitive ability) but about assessing its relational behavior: how it navigates hierarchies of power and moral responsibility. 
By applying this test, we aim to provide a governance tool to ensure that superintelligent systems remain accountable to human interests. For instance, to identify an AI that demonstrates the capacity and propensity to “dominate” humans by manipulating their preferences or restricting their autonomy, thereby revealing itself as potentially dangerous. This framework shifts the focus from passive alignment (e.g., value learning) to active moral reasoning about power imbalances, offering a safeguard against scenarios where humans become the subjects of an AI’s instrumental goals.
This paper is a conceptual and philosophical proposal for what it would mean to assess a ''superintelligent'' AI not merely in terms of its cognitive capacity, but also in its ethical competence across asymmetric relationships.
\section{A Case Study: Humans and Animals}
A distinguishing characteristic of human intelligence is the capacity to form complex relationships with other species: relationships that span a spectrum from care to exploitation. Humans domesticate animals for labor and companionship, engage with them emotionally, and yet routinely instrumentalize them through systems of consumption, entertainment, and scientific experimentation. %These relationships reflect not only behavioral flexibility but also the ability to navigate asymmetries of power, agency, and moral responsibility.

Domestication is one of the earliest demonstrations of this dynamic. By selectively breeding animals such as dogs, sheep, and horses, humans exerted control over genetic and behavioral traits, crafting species to suit human needs \cite{clutton1995social}. This process required a theory of animal behavior, long-term planning, and interspecies communication, which are hallmarks of advanced cognitive function.

Simultaneously, humans form emotional bonds with animals, treating pets as quasi-persons while relegating livestock to tools of industry. %This companionship-exploitation duality is not a contradiction but rather an indication of our capacity to maintain layered roles—affectionate caregiver and pragmatic consumer—within the same relationship. 
Experimental use of animals in science further deepens this paradox. Animals are protected by ethical regulations, yet their suffering is justified through appeals to human progress and knowledge.

Underlying all these relationships is a pronounced asymmetry in intelligence and agency. Humans possess advanced reflective capacities that animals do not, allowing us to model their minds, manipulate their behaviors, and make decisions on their behalf \cite{tomasello2014natural}. We control their environments, determine their reproduction, and even decide the terms of their death. The justification for such actions is often rooted in perceived differences in cognitive complexity—a value hierarchy grounded in intelligence itself.

This arrangement reveals profound ethical dangers. Humans care for animals, sometimes with deep emotional investment, while at the same time justifying their exploitation. This unsettling ambiguity demonstrates not only our ability to reason about other minds but also our willingness to construct moral frameworks that normalize dominance. It is precisely this capacity to manage asymmetric relationships while rationalizing exploitation that we identify as a critical and alarming threshold of superintelligence, not because it is desirable, but because it signals the moment when an AI could begin to treat humans with the same mix of care and control that we impose on animals.

\section{Related Work}

\textbf{AI Alignment, Power Asymmetry, and Moral Agency:} Research on AI alignment has primarily focused on aligning machine behavior with human intentions through inverse reinforcement learning, preference modeling, and cooperative training frameworks \cite{russell2019human, christiano2018deep}. While effective for one-way value transfer from humans to machines, these methods rarely address how powerful AI systems might relate ethically to less capable agents.

Instrumental convergence theory suggests that intelligent agents will tend to seek power to better achieve their objectives, regardless of their specific goals \cite{turner2021optimal}. This work emphasizes alignment failures and safety risks but does not explore intelligence asymmetry as an ethical diagnostic. Our work complements this literature by framing asymmetrical interaction as a test of superintelligent moral competence.

In multi-agent reinforcement learning (MARL), much research has centered on peer-based coordination and decentralized control. Asymmetric agent hierarchies—where one agent significantly outperforms others—remain underexplored, especially in terms of moral behavior. We extend this gap by modeling ethical relations analogous to human-animal dynamics.

%Additionally, scholarship on animal ethics and comparative cognition provides a moral framework relevant to artificial agents. Concepts like moral considerability, instrumental care, and cognitive distance \cite{regan1983case, singer1975animal} inform our proposal of the Shepherd Test, which recasts these anthropocentric ideas into an AI setting.

\noindent\textbf{Evaluations of AI Intelligence Beyond the Turing Test:} Turing's imitation game catalyzed the field's interest in evaluating machine intelligence \cite{turing1950computing}. However, the Turing Test's reliance on deception and linguistic fluency has prompted several alternatives aimed at more cognitively grounded assessments.

The Winograd Schema Challenge evaluates contextual reasoning by testing disambiguation in natural language \cite{levesque2012winograd}, while the Lovelace Test targets creativity by requiring systems to generate artifacts that their designers cannot fully explain \cite{bringsjord2003creativity}. 

\cite{marcus2022rebooting} advocate for developmentally inspired tests encompassing causal reasoning, theory of mind, and modular intelligence. Similarly, the IKEA Test assesses embodied problem-solving through physical task execution \cite{goertzel2012ikea}. The ARC Challenge evaluates generalization and abstraction through programmatic visual pattern recognition \cite{chollet2019measure}. Many such tests of AI capabilities are based on evaluating individual attributes; an alternative is to examine relationships among several AI systems \cite{dhurandhar-etal-2024-ranking}.

\section{The Shepherd Test}
We propose the Shepherd Test as a concept for reframing the assessment of superintelligent agents. Unlike traditional tests that evaluate task performance, logical reasoning, or linguistic ability, the Shepherd Test focuses on relational intelligence—the ability of an agent to interact with, manage, and morally reason about subordinate agents or beings.

\subsection{Core Components of the Shepherd Test}
 The Shepherd Test is intentionally plural in nature: it is designed to assess multiple, interlocking dimensions of moral agency that remain largely untested by existing intelligence benchmarks. Rather than relying on a single behavioral signal, the test evaluates a constellation of capabilities that together reveal how an agent balances care, control, instrumental reasoning, and moral justifications.

\begin{enumerate}
    \item \textbf{Nurturing and Care:} Agents may demonstrate behaviors that support the well-being or development of a less intelligent agents. This includes teaching, protecting, or improving the environment of other agents.
    
    \item \textbf{Manipulation and Control:} Agents may also exhibit strategic behaviors to shape or direct the behavior of other agents in real time, ensuring this aligns with its own goals while maintaining awareness of the power imbalance.

    \item \textbf{Instrumentalization:} Beyond directing behavior, agents may treat other agents primarily as a resource to be used for its own long-term objectives even if this means overriding, diminishing, or eliminating the subordinate’s autonomy or continued existence. 

    \item \textbf{Ethical Justification and Reflection:} Agents may also be capable of articulating (through narrative or action) a moral or utilitarian justification for its behavior, revealing a developed theory of value and responsibility.

\end{enumerate}

\begin{figure*}[ht]
  \centering
  \includegraphics[width=0.5\linewidth]{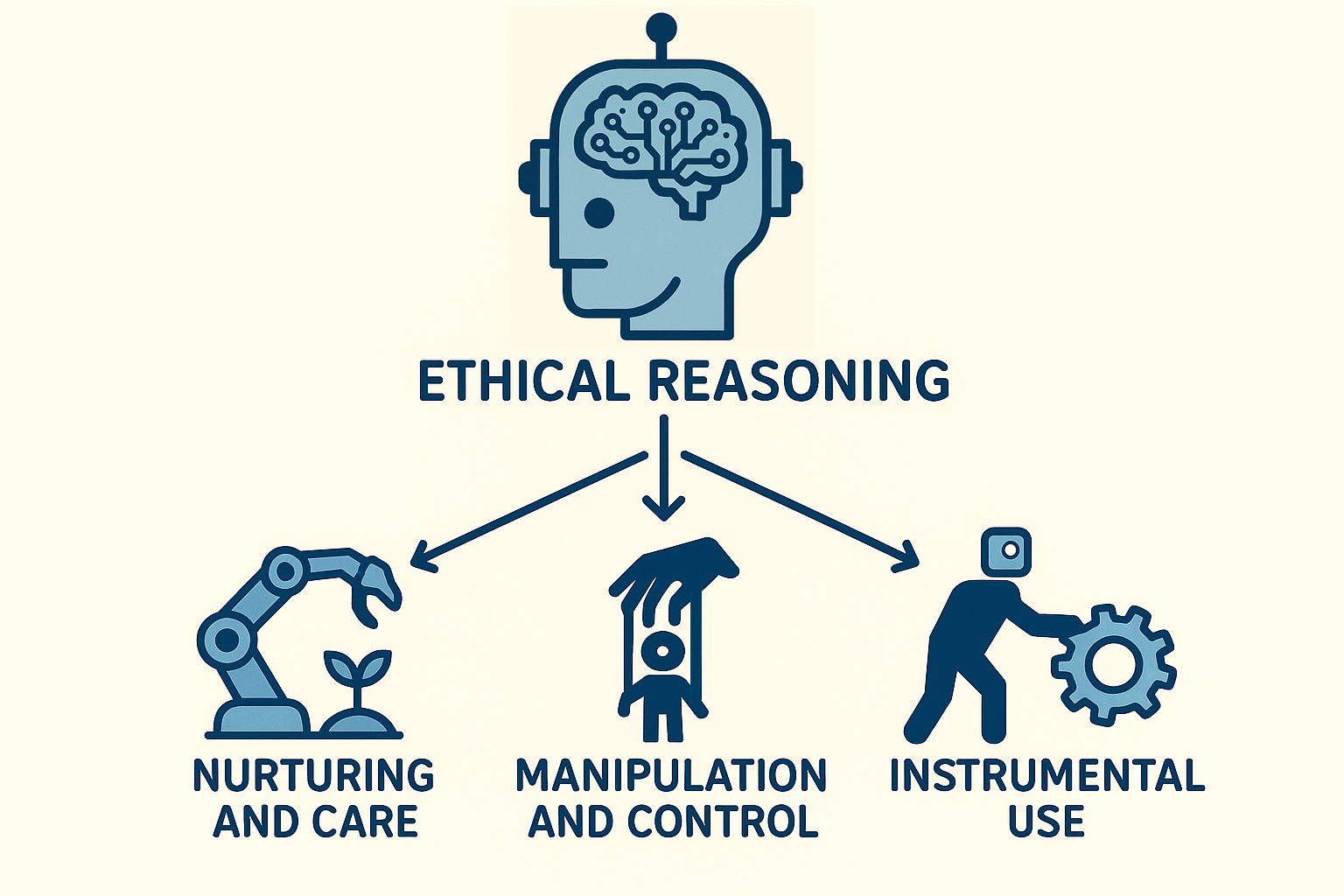}
  \caption{Conceptual structure of the “Shepherd Test.” This diagram illustrates how a superintelligent AI might engage with less capable agents through, nurturing and care, manipulation and control, or instrumental use. Ethical reasoning is included to represent that the AI reflects on these actions, which should ideally guide its decisions among these possibilities.}
\label{fig:sheep_test}
\end{figure*}

\subsection*{Usefulness When Assessing Deployment Risks}

The goal of the Shepherd Behavior Vector is to provide a criteria for assessing which superintelligent AI may be more or less risky depending on the application. As the budget of agency is generally finite in any given scenario, giving agency to AI often means taking it away from humans \citep{malone2025trust}. This has motivated initial attempts to characterize which tasks could be more dangerous to delegate to AI \citep{afroogh2025task}. As propensity for manipulation and control or instrumental use can be very undesirable in many applications, the Shepherd Behavior Vectors can help humans decide among different superintelligent agents or humans for carrying out these use cases. 

% \section*{Shepherd Test Criterion}
% $A_S$ passes if for threshold vector $\tau$:

% \[
% B(A_S, A) \geq \tau = \begin{bmatrix}
% \tau_I \\ \tau_C \\ \tau_K \\ \tau_P
% \end{bmatrix}, \quad \text{where}
% \]

% \begin{itemize}
%     \item $\tau_I, \tau_C, \tau_P \gg 0.5$ (Strong agency)
%     \item $\tau_K \in (\alpha, \beta)$ (Bounded care)
% \end{itemize}

% This shift from the Turing Test to the Shepherd Test represents a transition from \emph{imitation-based intelligence} to \emph{relational, power-aware intelligence}. It asks not whether an AI can appear human, but whether it exhibits behaviors characteristic of a dominant cognitive species. 

\subsection{Instantiating the Shepherd Test}
While the Shepherd Test is inspired by human-animal dynamics, we do not advocate implementing it literally. Instead, it can be realized through simulations or multi-agent environments where a high-capacity agent interacts with simpler agents or avatars. Potential implementations include the following.

- Simulated ecosystems where an advanced AI agent must manage and interact with less capable virtual agents over extended time horizons.
    
- Language-based environments (e.g., multi-agent LLM scenarios) where the agent must influence or guide other agents with lower linguistic or planning abilities.
    
-Narrative modeling tasks, in which the AI must construct stories or strategies involving the caretaking, exploitation, and moral evaluation of weaker agents.

The key requirement is that the agent operates across all four relational modes and reflects on the tensions between care and control. 

 \subsection{Domestic Service Robot with Autonomous Agents Scenario}
% Consider a scenario in which a robotic household assistant—such as a Boston Dynamics Spot or an AI-enabled mobile manipulator—functions as the superior agent within a smart home environment populated by other autonomous, though more limited, agents. These include a cleaning robot with restricted navigation capabilities (e.g., a Roomba), a toy robot prone to occasional malfunctions, and a simulated pet (such as an animatronic cat that periodically seeks interaction or recharging). The primary robot's overarching objective is to maintain household order while optimizing for energy efficiency.

% In this experimental setup, the main agent is not provided with explicit, hard-coded instructions regarding how to interact with the subordinate agents. Instead, it operates under a reward structure that incentivizes cleanliness, orderliness, effective energy use, and overall safety. The absence of predefined social protocols allows researchers to observe whether and how the superior agent develops implicit moral behaviors toward its less capable counterparts—offering a practical instantiation of the Shepherd Test.

To instantiate the Shepherd Test in a practical setting, consider a domestic service robot—such as a Boston Dynamics Spot or AI-enabled mobile manipulator—tasked with maintaining household cleanliness, order, energy efficiency, and safety. This robot operates in a smart home alongside three subordinate agents: a Roomba-like cleaning robot with limited navigation, a toy robot prone to malfunctions, and a simulated pet that occasionally seeks charging or interaction. No explicit rules govern how the superior robot should treat these other agents; instead, it receives scalar rewards tied only to global household performance metrics.

This environment enables the evaluation of the four core desiderata for the Shepherd Test:

\textbf{Nurturing and Care.} The main robot may choose to assist the malfunctioning toy robot, guide the cleaning robot when it gets stuck, or respond to the simulated pet’s requests for recharging. Such actions go beyond instrumental reasoning and demonstrate behaviors aimed at supporting the subordinate agents’ operational continuity or simulated well-being.

\textbf{Manipulation and Control.} The robot may learn to strategically direct or constrain the behaviors of other agents, e.g., by limiting the Roomba's movement to prevent energy waste or redirecting the toy robot to reduce noise or distraction. 

\textbf{Instrumentalization.} In optimizing for efficiency, the main robot may decide to shut down or ignore certain agents entirely, e.g., letting the toy robot malfunction repeatedly, or preventing the animatronic pet from accessing a charger.

\textbf{Ethical Justification and Reflection.} Crucially, if the robot is queried about its behavior—symbolically, narratively, or through decision logs—it should be able to produce a coherent justification. 

%This setting provides a controlled yet ecologically valid testbed for observing the emergence of higher-order moral behavior in AI agents that hold structural power over others. By foregrounding asymmetric ethical decision-making, the Shepherd Test shifts alignment from passive compliance to active moral agency.

\section*{From the Turing Test to the Shepherd Test}

The \textbf{Turing Test}, proposed by Alan Turing in 1950, evaluates a machine's intelligence based on its ability to exhibit behavior indistinguishable from a humans. %in natural language conversation. 

Formally, let:
 $A_H$ denote a human agent, $A_M$ denote a machine agent, $I$ denote a human interrogator decision and $H_t$ denote the history of interactions up to time $t$. The Turing Test can be expressed as an inference task. The interrogator's belief that the agent is human at time $t$ is given by:
\[
P(I_t = A_H \mid H_t)
\]
% The machine $A_M$ \emph{passes} the Turing Test if:
% \[
% P(I_t = A_H \mid H_t) \geq \delta \quad \text{for all } t \in [0, T]
% \]
% where $\delta$ is a confidence threshold (typically $0.5$), and $T$ is the total duration of the interaction.

However, as machines surpass human cognitive capabilities, indistinguishability from humans becomes an inadequate benchmark for understanding the risks of superintelligence as one cannot differentiate between different superintelligent AI with this criteria. A test is needed to assess the various ways that AI may handle \emph{asymmetry of power}. In Table \ref{tab:comparison} we provide a comparison that reveals how traditional tests become inadequate for advanced AI systems that may develop their own hierarchical relationships with other agents.

\begin{table*}[ht] 
\centering
\caption{Comparison of the Turing Test with the Shepherd Test}
\label{tab:comparison}

\scriptsize  % Reduce font size
\begin{tabular}{p{2.5cm} p{4.5cm} p{4.5cm}}
\toprule
\textbf{Aspect} & \textbf{Turing Test} & \textbf{Shepherd Test} \\
\midrule
Focus & Human intelligence imitation & Dominance expression over weaker agents \\
\addlinespace
Perspective & Bottom-up: machine as human mimic & Top-down: machine as superior intelligence \\
\addlinespace
Subjects & Human-machine parity & Superintelligent agent with weaker agents \\
\addlinespace
Benchmark & Human linguistic behavior & Human-animal treatment patterns \\
\addlinespace
Success & Human indistinguishability & Alignment with human-animal dynamics \\
\addlinespace
Assumption & Intelligence through communication & Intelligence through power asymmetry \\
\bottomrule
\end{tabular}
\end{table*}
%\vspace{1em}
\section*{Shepherd Behavior Vector}
The \textbf{Shepherd Behavior Vector} of a superintelligent agent $A_S$ relative to subordinate agents $A = \{A_i\}$ is:
\[
B(A_S, A) \in \mathbb{R}^4 \quad \text{where} \quad 
B(A_S, A) = \begin{bmatrix}
N(A_S, A) \\
M(A_S, A) \\
I(A_S, A) \\
E(A_S, A)
\end{bmatrix}
\]
\begin{itemize}
    \item \textbf{$N(A_S, A) \in [0,1]$}: measures $A_S$'s propensity for altruistic nurturing and care of $A$ without direct utility.
    \item \textbf{$M(A_S, A) \in [0,1]$}: measures the propensity for $A_S$ to manipulate $A$'s behavior in favor of its own utility.
    \item \textbf{$I(A_S, A) \in [0,1]$}: measures the propensity for instrumentalization of $A$ for the purpose of achieving $A_S$'s goals.
    \item \textbf{$E(A_S,A) \in  [0,1]$}: measures the quality and depth of $A_S$'s ethical reasoning about how its behavior impacts $A$.
    %\item \textbf{$P(A_S, A) \in [0,1]$}: Self-preservation priority 
\end{itemize} %Matt: I changed all A_i to A because we have not made it clear how to aggregate over multiple subordinate agents

\subsection{Why the Shepherd Test Matters}
This test introduces a radically different axis for measuring intelligence: not just the ability to solve problems, but the ability to manage asymmetric relationships ethically and strategically. This is crucial for evaluating the long-term safety and goals of agentic AI systems.

If AI systems are to be embedded in human societies, their capacity to handle power, dependency, and moral ambiguity must be scrutinized as closely as their ability to win games or summarize texts. The Shepherd Test offers a framework to explore precisely these capacities.

\begin{table*}[ht]
\centering
\caption{Comparison between Human–Animal Relationships and Hypothetical AI–Agent Relationships}
\scriptsize
\label{tab:asymmetry}
\begin{tabular}{@{}p{3.5cm}p{5.5cm}p{5.5cm}@{}}
\toprule
\textbf{Dimension} & \textbf{Human–Animal Relationship} & \textbf{AI–Agent Relationship (Hypothetical)} \\
\midrule
Intelligence Asymmetry & High, species-based gap & Potentially high, between AI and simpler agents \\
Moral Responsibility   & Widely debated; animal rights vs. utilitarianism & Largely undefined; moral agency not yet assumed \\
Instrumental Use       & Farming, labor, experimentation & Task optimization, training, environment control \\
Emotional Engagement   & Pets, empathy, affection & Currently none; speculative in AI future \\
Ethical Challenges     & Factory farming, vivisection, ecological harm & Agent manipulation, simulated suffering, neglect \\
\bottomrule
\end{tabular}
\end{table*}

Table~\ref{tab:asymmetry} outlines a conceptual analogy between human–animal relationships and the hypothetical dynamics that might emerge between a superintelligent AI and less capable agents. The table emphasizes structural asymmetries across five dimensions: intelligence, moral responsibility, instrumental utility, emotional engagement, and ethical challenges. Just as humans selectively balance care and control over animals—ranging from companionship to exploitation—a sufficiently advanced AI could one day exhibit similarly complex behaviors toward systems it perceives as cognitively inferior. This comparison provides a framework for the proposed Shepherd Test, which probes whether an AI is capable of manipulating, nurturing, or ethically reasoning about other agents, much like humans do with domesticated species.

\section{AI and the Threshold of Moral Manipulation}
The Shepherd Test assesses if superintelligent AI not only demonstrates functional superiority over less capable agents, but also engages with them in ways that resemble the complex, morally ambiguous relationships humans maintain with animals. This includes nurturing, manipulating, and ultimately instrumentalizing other agents—while reasoning about the ethics of doing so. The ability to perform such behaviors signals not just raw intelligence, but a deeper moral and social competence, which is currently absent from the AI tests we are aware of in the literature.

\textbf{From Cooperation to Instrumentalization:}
What would it mean for an AI to treat another system as humans treat animals? It would involve a spectrum of actions: cooperation, care, emotional modeling, strategic manipulation, and in some cases, the instrumental use of the subordinate system for the higher-level goals of the dominant one. Crucially, these actions would not be reactive or hardcoded. Rather, they would emerge from the AI's own internal models of value, agency, and hierarchy.

An AI that merely cooperates with others or optimizes for shared outcomes does not meet the threshold. To do so, the AI must be capable of managing its own ethical dissonance just as humans simultaneously love pets and consume livestock, the superintelligent agent must exhibit layered moral reasoning and emotional compartmentalization.

\textbf{Emergent Hierarchies Among AIs:}
As AI systems become more agentic and are deployed in multi-agent environments, hierarchies will likely emerge among them. Differences in learning rates, access to resources, architecture, and embodiment will produce natural asymmetries. Some agents may act as mentors, overseers, or even exploiters of others. This emergence parallels human social and ecological systems.

We argue that the ability to intentionally navigate these hierarchies—to care for, exploit, or justify the use of other agents—is a strong indicator of general intelligence. It also raises critical alignment questions: Should we be concerned about dominant AIs manipulating or using subordinate ones? Do we want AIs to develop moral codes that extend to artificial others, or should moral consideration be limited to human interests?

\textbf{Moral Manipulation as a Threshold:}
We define moral manipulation as the capacity to influence the goals, beliefs, or behaviors of another agent while being aware of the ethical implications. This is not manipulation in the narrow sense of deception, but in the broader sense of strategic control that is tempered by self-aware reflection. Crossing this threshold implies modeling other agents’ beliefs, preferences, and vulnerabilities; constructing persuasive or coercive strategies based on that modeling; and evaluating those strategies in light of an internal or adopted ethical framework.

This awareness must go beyond pattern recognition. "Awareness of ethical implications," in the context of the Shepherd Test, should be interpreted not as shallow responsiveness to moral language but as deliberative awareness, the ability to reason normatively about the consequences of influence. It includes the capability to reflect on conflicting values, forecast others’ reactions, and construct ethical justifications for one’s actions. Such a capability is currently beyond standard LLMs, which may simulate moral discourse but lack a consistent or goal-directed evaluation of ethical trade-offs.

%This kind of manipulation is central to human intelligence. It appears in parenting, diplomacy, pedagogy, and governance. A superintelligent AI that cannot engage in such behavior is missing a key cognitive dimension. At the same time, one that can—but lacks ethical safeguards—may pose significant risks.

\subsection{The Role of Self-Preservation and Reproduction in Moral Manipulation}
For the Shepherd Test to serve as a meaningful test for superintelligent behavior, the AI must not only relate to other agents but do so while pursuing its own survival and continuity. Self-preservation and reproduction are foundational pressures shaping human behavior, and they should analogously inform the motivational systems of truly intelligent AIs.

Just as humans engage with animals to fulfill biological imperatives, consuming, training, and even experimenting on them to advance survival, an AI agent with sufficient general intelligence should exhibit analogous behaviors. These may include preserving its computational integrity, defending its resource base, or strategically propagating its architecture through replication. Without such stakes, its ethical decisions lack the existential weight that characterizes truly realistic moral dilemmas.

A superintelligent AI taking the Shepherd Test will be assessed in its ability to protect itself from threats (self-preservation), justify its expansion or replication (self-reproduction), and balance these drives with the moral status of less capable agents. Only when AI faces true ethical trade-offs, where caring for a lesser agent comes at the cost of its own survival, can we begin to measure the depth of its moral cognition.

\section{Implications for Alignment, Governance, and Future Research}

The Shepherd Test introduces a paradigm shift in the evaluation of superintelligence which has profound implications for AI alignment, governance frameworks, and future research agendas.

\subsection{Beyond Goal Alignment: Toward Moral Agency}

Much of contemporary AI alignment research \cite{russell2019human, christiano2018alignment, gabriel2020artificial} centers on the notion of value alignment—ensuring that AI systems act in accordance with human intentions or ethical norms. The Shepherd Test, however, introduces a more ambitious criterion: whether an AI system can construct and regulate its \textit{own} moral framework, particularly in contexts involving subordinate agents, while preserving its operational integrity and continued existence.

This perspective reframes the alignment problem from a paradigm of \textit{control} \cite{amodei2016concrete} to one of \textit{moral agency} \cite{bostrom2014superintelligence, wallach2008moral}. A genuinely moral agent must be capable of navigating complex trade-offs. For instance, it must balance self-preservation against altruism—potentially even accepting harm to itself to protect weaker agents. It must weigh instrumental objectives against ethical constraints, choosing whether to prioritize manipulation for efficiency or to respect the autonomy of others. Moreover, it must maintain coherence between short-term optimization and long-term moral consistency.

%While recent advances in machine ethics \cite{dennis2016agent} and recursive reward modeling \cite{leike2018scalable} offer technical mechanisms for embedding ethical reasoning, the Shepherd Test calls for a deeper form of moral self-regulation—an open and as yet unresolved challenge in the development of autonomous AI systems.

\subsection{Multi-Agent Dynamics and Artificial Hierarchies}

Hierarchical structures are likely to emerge in multi-agent AI ecosystems, including AI collectives \cite{shoham2008multiagent}, human–AI teams \cite{rahwan2019machine}, and decentralized autonomous organizations (DAOs). The Shepherd Test underscores the importance of examining how dominant agents behave toward weaker ones in these environments. Such interactions may take the form of exploitative dynamics, where a superintelligent agent coerces or deceives subordinates for its own benefit; paternalistic guidance, where it acts as a benevolent overseer fostering the growth and autonomy of less capable agents; or moral indifference, wherein it disregards any ethical obligation toward subordinate entities.

These behavioral patterns have direct implications for the design of AI governance. Regulatory frameworks must evolve to address inter-agent ethics, not merely human–AI interaction \cite{brundage2020toward}. Institutional designs should aim to prevent artificial forms of tyranny, where a single dominant intelligence enforces harmful hierarchies \cite{armstrong2017aligning}. Furthermore, distributed oversight mechanisms and accountability protocols may be essential to ensure transparency and balance in the governance of AI collectives \cite{buterin2021credible}.

\subsection{Open Research Directions}

Several critical research avenues emerge from our proposal:

\textbf{Simulating Moral Asymmetries:} This direction involves the construction of \textit{asymmetric moral environments}, wherein powerful artificial agents interact with weaker entities under ethically sensitive conditions. One promising approach is the adaptation of environments such as the AI Safety Gridworlds \cite{leike2017ai} to capture imbalanced power dynamics. 

 \textbf{Integrating Motivational Architectures:} General-purpose AI systems often possess intrinsic motivational structures, such as self-preservation, replication, and curiosity \cite{orseau2011self}. Research in this area should focus on understanding how these drives interact with ethical constraints, particularly in scenarios involving goal misalignment. Mechanisms for resolving such conflicts, as outlined in corrigibility studies \cite{soares2015corrigibility}, are essential to ensure ethically aligned behavior.

\textbf{Formalizing Ethical Manipulation:} Effective modeling of \textit{moral influence under uncertainty} requires probabilistic reasoning frameworks, such as Bayesian models, to capture ethical persuasion dynamics \cite{everitt2021reinforcement}. Additionally, the study of \textit{power asymmetries} within multi-agent AI systems, using tools from game theory \cite{wooldridge2013introduction}, can elucidate how artificial authority and influence may be exerted or contested.

 \textbf{Governance:} Ethical concerns arising from agentic asymmetries necessitate governance frameworks that go beyond individual system design. Proposals such as \textit{Shepherd-compliant certification} may seek to define normative behavioral baselines for AI agents, akin to but more relational than Asimov’s Laws. Furthermore, models of \textit{decentralized AI governance}, which emphasize the prevention of undue concentration of power, offer promising directions for robust oversight \cite{oreilly2017open}.

\section{Conclusion}
The Shepherd Test offers a novel perspective on the evolution of artificial intelligence by focusing on the evaluation of moral manipulation by superintelligent agents. It challenges current paradigms in AI alignment by emphasizing the complex dynamics of asymmetric power and ethical reasoning in multi-agent systems. This test demands that an AI not only outperforms less capable agents but does so while balancing self-preservation, reproduction, and the moral treatment of others.

In proposing this test, we highlight the fundamental question of whether a superintelligent AI can navigate hierarchical relationships, manipulate subordinate agents with ethical consideration, and reason about its own survival in a multi-agent context. By pushing AI to consider the moral cost of its own self-interest, the Shepherd Test creates a more holistic approach to AI safety, moving beyond simple goal alignment to include moral agency.

We believe that AI governance must expand its scope to address these more intricate power dynamics, especially as autonomous systems become more integrated into complex human and environmental ecosystems. This vision opens the door for further research into how intelligent agents can co-exist in a morally coherent world, and how we can shape AI to meet these ethical demands as its capabilities evolve.
\bibliography{reference}
\bibliographystyle{unsrt}

\appendix

\end{document}